\documentclass[cameraready]{Interspeech}
\title{Segment-level Tree Search for Long Meeting Document Summarization}

\author[affiliation={1}]{Sangwon}{Ryu}
\author[affiliation={3}]{Heejin}{Do}
\author[affiliation={1}]{Jun}{Seo}
\author[affiliation={4}]{Daehui}{Kim}
\author[affiliation={5}]{Yunsu}{Kim}
\author[affiliation={1,2}]{Gary Geunbae}{Lee}
\author[affiliation={1,2}]{Jungseul~}{Ok}
\address{
    $^1$ GSAI, POSTECH \quad
    $^2$ CSE, POSTECH\\
    $^3$ ETH Zurich, ETH AI Center \quad
    $^4$ Agentic AI Lab, KT \quad
    $^5$ LILT
}

\email{\{ryusangwon, sjin4861, gblee, jungseul\}@postech.ac.kr, heejin.do@ai.ethz.ch, daehui.kim@kt.com, yunsu.kim@lilt.com}

\keywords{natural language generation, abstract spoken document summarization, tree search}

\usepackage{comment}
\usepackage{multirow}

\begin{document}

\maketitle

\begin{abstract}
% 1000자 제한
Meeting documents are challenging to summarize due to their length and complex conversational structure. 
Existing approaches typically adopt multi-stage pipelines that extract information prior to summarization; however, these approaches often suffer from cumulative error propagation without intermediate validation, a limitation further amplified by short and low-quality reference summaries.
We propose \underline{s}egment-level \underline{s}ummarization via Monte Carlo Tree \underline{S}earch (S3), a training-free framework that constructs a final summary by composing segment-level summary candidates. S3 partitions a long document into segments and generates multiple summary candidates per segment, forming nodes of a search tree. The best-scoring combination is selected via self-reward-guided tree search and refined into the final output. Despite using a 7B model, S3 achieves performance comparable to larger 72B models while producing length-appropriate summaries.

\end{abstract}

\iffalse
\begin{abstract}
    
Summarization of meeting documents is particularly challenging due to their length, complex conversational structure, and multi-speaker dynamics. 
Existing methods typically adopt multi-stage pipelines that extract information prior to summarization; however, these approaches often suffer from cumulative error propagation without intermediate validation, a limitation further amplified by short and low-quality reference summaries.
We propose \underline{s}egment-level \underline{s}ummarization via Monte Carlo Tree \underline{S}earch (S3), a training-free framework that constructs a final summary by composing segment-level summary candidates. S3 partitions a long document into segments and generates multiple summary candidates per segment, forming nodes of a search tree. A self-reward guided tree search then selects an optimal combination, which is refined into a final output. Despite using a 7B-scale LLM, S3 achieves performance comparable to larger 72B-scale models while producing length-appropriate and structurally coherent summaries.

\end{abstract}
\fi

\begin{figure*}[ht]
\centering
\includegraphics[width=0.85\textwidth]{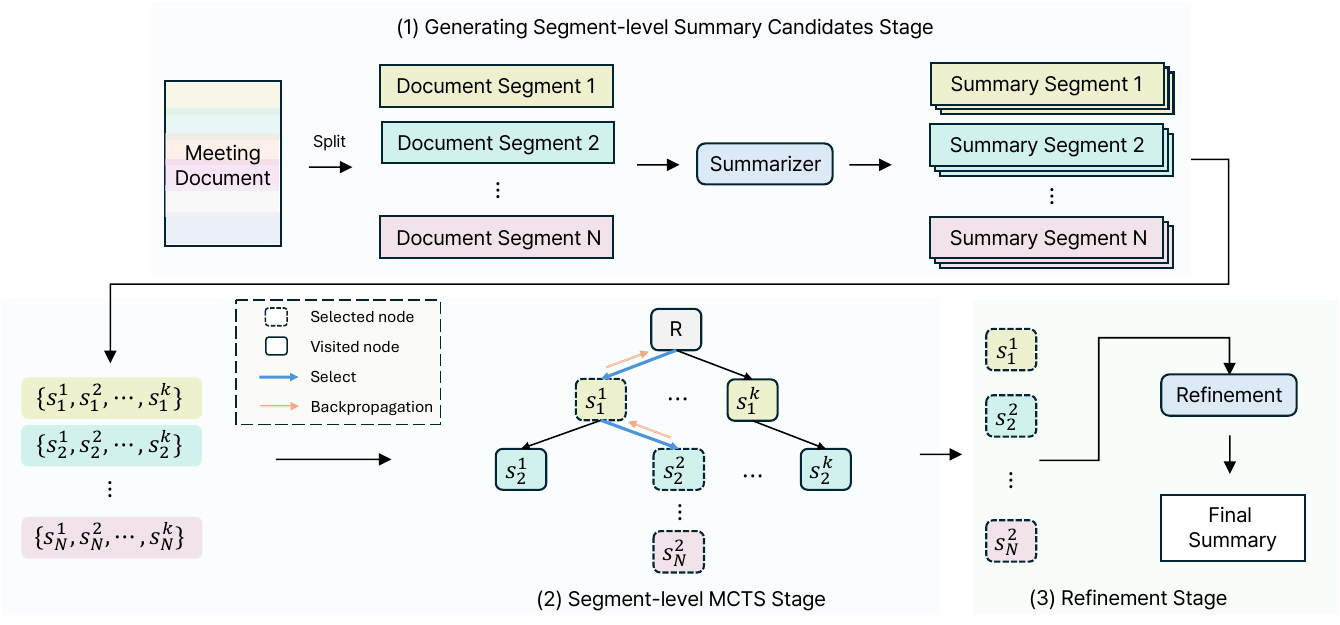}
\caption{Overall framework of S3. (1) The meeting document is first divided into segments, and multiple summary candidates are generated for each segment. (2) These candidates form a search tree, where segment-level combinations are explored using MCTS with self-evaluation. (3) The selected segment-level summaries are concatenated and refined to produce the final coherent summary.}
\label{fig: main}
\end{figure*}

\section{Introduction}

Meeting summaries are widely used to help both attendees and non-attendees understand and recall prior discussions. In practice, overly short summaries are often insufficient, as they fail to convey the context and detailed information underlying key decisions and outcomes. However, existing meeting summarization datasets predominantly provide short reference summaries, implicitly encouraging models to generate overly compressed outputs and offering limited supervision for generating comprehensive summaries. This mismatch between real-world needs and available data hinders the development of systems capable of producing informative summaries. Moreover, important decisions and rationales in long meetings are often dispersed across unstructured conversational transcripts, where key information is sparsely distributed rather than concentrated, making effective summarization challenging.

Existing approaches to long meeting summarization typically rely on multi-stage pipelines that extract salient information prior to summarization~\cite{galley-etal-2003-discourse, zhang-etal-2022-summn, mao-etal-2022-dyle, vig-etal-2022-exploring, kim-etal-2023-explainmeetsum, ou-lapata-2025-context} due to the limited context window of encoder-decoder models. Although this approach alleviates context length constraints, such pipelines often suffer from cumulative error propagation, as mistakes introduced at early stages cannot be corrected in later steps due to the lack of intermediate validation. This problem is amplified in meetings, where conversational noise and frequent topic shifts increase the impact of early-stage errors.

Recent large language models (LLMs) have substantially expanded context windows, with context lengths exceeding 100K tokens~\cite{grattafiori2024llama3herdmodels, qwen2.5, openai2024gpt4technicalreport}. While longer context windows make end-to-end processing of entire meeting transcripts feasible, our preliminary analysis suggests that increased input length alone is insufficient to effectively handle globally dispersed information in long meetings.

To address these challenges, we propose segment-level summarization via Monte Carlo Tree Search (S3), a training-free framework that constructs a final summary by composing segment-level summary candidates. S3 first partitions a long meeting document into segments and generates multiple candidate summaries for each segment, forming the nodes of a search tree. A self-reward–guided Monte Carlo Tree Search (MCTS) then selects an optimal combination of segment-level summaries. Finally, a refinement step removes redundant or generic expressions across segments to enhance global coherence in the final assembled summary. By explicitly reasoning over combinations of segment-level summaries, S3 captures globally distributed important information while maintaining structural consistency and appropriate summary length.

We evaluate S3 on the QMSum~\cite{zhong-etal-2021-qmsum}, meeting summarization benchmark, using LLMs capable of processing long inputs. Summary quality is assessed using G-Eval~\cite{liu-etal-2023-g} across multiple dimensions, including \textit{coherence}, \textit{consistency}, \textit{fluency}, and \textit{relevance}. In addition, we analyze model performance across different input-length bins. Experimental results demonstrate that S3 with a 7B model substantially outperforms baseline models with the same backbone and achieves performance comparable to that of much larger 72B model. Notably, S3 consistently achieves the strongest performance across all input-length bins, with the largest gap on the longest documents.

\section{Related work}
The unstructured and conversational nature of meeting transcripts poses significant challenges for summarization, often resulting in missing key information, redundancy, hallucinations, and incoherent outputs~\cite{tang-etal-2022-confit, rennard-etal-2023-abstractive, zou-etal-2023-towards, deutsch-roth-2023-incorporating, kirstein-etal-2024-whats, ryu24_interspeech}.
%Summarizing long meeting documents is challenging due to their unstructured and conversational nature, often leading to issues such as missing important information, redundancy, hallucination, and incoherent or poorly organized summaries~\cite{kirstein-etal-2024-whats, ryu24_interspeech}. 
To handle long inputs, prior work has primarily relied on multi-stage strategies such as extract-then-generate~\cite{mao-etal-2022-dyle} or divide-and-conquer~\cite{zhang-etal-2022-summn, vig-etal-2022-exploring}. However, these approaches are prone to cumulative error propagation and often depend on short, low-quality reference summaries for supervision. Moreover, evaluation has largely relied on ROUGE~\cite{lin-2004-rouge}, which may not adequately capture summary quality in long-form meeting settings \cite{rennard-etal-2023-abstractive, gong25c_interspeech}. %while evaluation is typically limited to ROUGE. 
Recent advances in long-context language modeling~\cite{grattafiori2024llama3herdmodels, qwen2.5, openai2024gpt4technicalreport}, supporting input-length exceeding 100K tokens, enable more fine-grained modeling and evaluation of extended conversations. Building on these developments, we explore the potential of LLMs for long meeting summarization and propose S3 framework that constructs summaries without relying on reference summaries.
%With recent advances in tokenization enabling input lengths exceeding 100K tokens, more fine-grained summarization and evaluation have become feasible. We explore the performance of LLMs on long meeting document summarization and propose S3, a self-reward-based Monte Carlo Tree Search (MCTS) framework that handles long meeting documents without relying on reference summaries.

\section{Method: S3}

We propose \underline{s}egment-level \underline{s}ummarization via Monte Carlo Tree \underline{S}earch (S3) to generate informative summaries for long meeting documents in which important information is distributed across the entire conversation. S3 first divides a long document into fixed-length segments and generates multiple summary candidates for each segment through sampling. 
To improve inference efficiency in MCTS-based summarization~\cite{ryu2025adaptive}, all candidate summaries are generated in advance and used to construct the search tree offline, where each segment corresponds to a depth level.
% Each segment corresponds to a depth level in a search tree, and we construct the tree structure offline using these candidate summaries. 
We then apply self-reward-based MCTS to identify an optimal combination of segment-level summaries and concatenate them into a single draft summary. Finally, a refinement step is performed to remove redundancy across segments and improve coherence, producing the final summary.
The overall framework is illustrated in Figure~\ref{fig: main}.

\begin{table*}[ht]
\centering
\caption{Comparison of models on on the QMSum dataset. \textbf{Bold} indicates the best performance within each baseline model.}
\scalebox{0.86}
{\begin{tabular}{l|c|c|ccccc|c}
\toprule
\textbf{Model} & \textbf{Granularity} & \textbf{\# of Params} & \textbf{Coherence} & \textbf{Consistency} & \textbf{Fluency} & \textbf{Relevance} & \textbf{Average} & \textbf{ROUGE-1}\\
\midrule
Reference summary & - & - & 2.62 & 3.00 & 4.02 & 2.87 & 3.13 & - \\
% \midrule
% SEGENC (BART$_{\text{large}}$)$^{*}$ & Summary-level & 406M & - & - & - & - & - & -  \\
% \midrule
% Llama-3.2-Instruct & Summary-level & 3B & 3.78 & 3.86 & 4.70 & 4.05 & 4.10 & 0.220  \\
% \quad w/ Refine & Summary-level & 3B & 3.98 & 3.95 & \textbf{4.95} & 4.22 & 4.28 & 0.227  \\
% \quad w/ S2 + Refine & Segment-level & 3B & \textbf{4.11} & \textbf{4.19} & 4.92 & 4.41 & \textbf{4.41} & 0.147  \\
% \quad w/ \textbf{S3} & Segment-level & 3B & 4.10 & 4.12 & 4.92 & \textbf{4.47} & 4.40 & 0.142  \\
\midrule
Qwen2.5-Instruct & Summary-level & 7B & 4.01 & 4.22 & 4.90 & 4.36 & 4.37 & 0.287 \\
\quad w/ Refine & Summary-level & 7B & 4.00 & 4.18 & 4.93 & 4.34 & 4.36 & 0.285 \\
\quad w/ S2 & Segment-level & 7B & 4.10 & 4.33 & 4.91 & 4.55 & 4.47 & 0.174 \\
\quad w/ \textbf{S3} & Segment-level & 7B & \textbf{4.22} & \textbf{4.43} & \textbf{4.94} & \textbf{4.65} & \textbf{4.56} & 0.148 \\
\midrule
Gemma-3 & Summary-level & 12B & 4.07 & 4.31 & 4.92 & 4.41 & 4.43 & 0.254  \\
\quad w/ Refine & Summary-level & 12B & 4.14 & 4.27 & 4.96 & 4.48 & 4.46 & 0.242 \\
\quad w/ S2 & Segment-level & 12B & 4.41 & 4.66 & 4.97 & \textbf{4.83} & 4.72 & 0.152 \\
\quad w/ \textbf{S3} & Segment-level & 12B & \textbf{4.47} & \textbf{4.68} & \textbf{4.98} & \textbf{4.83} & \textbf{4.74} & 0.151 \\
\midrule
Qwen2.5-Instruct & Summary-level & 72B & 4.19 & 4.43 & 4.95 & 4.58 & 4.54 & 0.256 \\
% \quad w/ Refine & 72B & 4.17 & 4.36 & \textbf{4.97} & 4.57 & 4.52 & 0.249 \\
\bottomrule
\end{tabular}}
\label{tab: main_result}
\end{table*}

\subsection{Segment-level Sampling}

% We first divide a long meeting document into overlapping segments using a fixed window size of 2048 tokens with a stride of 256 tokens. 
We divide a long meeting document into overlapping segments using a sliding-window strategy with window size $w$ and stride $r$. 
This produces a sequence of text segments $\{t_1, t_2, \cdots, t_n\}$. For each segment $t_i$, we generate $k$ candidate summaries $\{s_i^1, s_i^2, \cdots, s_i^k\}$, resulting in the candidate set 
$\{\{s_1^1, \cdots, s_1^k\}, \{s_2^1, \cdots, s_2^k\}, \cdots, \{s_n^1, \cdots, s_n^k\}\}$. Candidate summaries are generated via nucleus sampling~\cite{holtzman2019curious} or diverse beam search (DBS)~\cite{vijayakumar2018diversebeamsearchdecoding} to encourage diversity across segment-level outputs. Since each candidate summary is generated from a localized document segment rather than the entire long document, the model can focus on a narrower context and produce more reliable and detailed summaries~\cite{do-etal-2025-multi}. The use of overlapping segments further mitigates issues such as coreference resolution errors and missing information near segment boundaries~\cite{vig-etal-2022-exploring}, helping preserve important content that spans consecutive segments.

\subsection{MCTS}

After generating $k$ candidate summaries for each document segment, we construct a search tree by organizing the segment-level summaries according to their corresponding depth. Each depth $d$ in the tree corresponds to a document segment $t_d$, and the available actions at that depth are the pre-generated summary candidates $\{s_d^1, s_d^2, \cdots, s_d^k\}$. LLM serves as the policy $\pi$, and each action corresponds to selecting one candidate summary segment at the current depth. S3 sequentially selects one summary candidate from $\{s_d^1, s_d^2, \cdots, s_d^k\}$ at each depth $d$, gradually constructing a complete summary path from the root to a leaf node. The depth of the tree is set to the number of document segments $n$, and the width at each depth is determined by the number of sampled candidates $k$. 

\subsubsection{Selection}

During the selection phase, starting from the root node, we recursively select child nodes according to the UCT~\cite{Kocsis2006} criterion until reaching a node that is not fully expanded or corresponds to a terminal state. At each state $s$, we select an action $a$ by maximizing:
\[
a^* = \arg\max_{a \in \mathcal{A}(s)}
\left(
Q(s,a) + c \sqrt{\frac{\log (N(s)+1)}{N(s,a)+1}}
\right),
\]

where $N(s)$ denotes the visit count of state $s$, $N(s,a)$ is the number of times action $a$ has been selected from $s$, and $c$ is an exploration constant that balances exploration and exploitation. The term $Q(s,a)$ represents the average accumulated reward obtained by taking action $a$ at state $s$.

\subsubsection{Expansion}
When a leaf node is reached during the selection phase, we expand the node by attaching its corresponding segment-level summary candidates as child nodes. Specifically, at depth $d$, we retrieve the pre-generated candidate set $\{s_d^1, s_d^2, \cdots, s_d^k\}$ and add each candidate as a separate child node. Each child node represents selecting one summary candidate for the current document segment, thereby extending the partial summary path by one depth level. This expansion process continues until the tree reaches depth $n$, which corresponds to the total number of document segments.

\subsubsection{Evaluation}
Once a complete summary path $\{s_1, s_2, \cdots, s_n\}$ is formed by selecting a candidate at the final depth, the full summary is obtained by concatenating the selected segment-level summaries. The constructed summary is then self-evaluated to compute a scalar reward $z$. Specifically, we evaluate the summary across four commonly used dimensions in summarization~\cite{liu-etal-2023-g, ryu-etal-2024-multi,fabbri2021summeval}: \textit{Coherence}, \textit{Consistency}, \textit{Fluency}, and \textit{Relevance}. Each dimension is rated on a Likert scale from 1 to 5. The scores are linearly normalized to the range $[-1, 1]$ and averaged to produce the final reward, defined as:

\[
\tilde{r}_i = 2 \cdot \frac{r_i - 1}{4} - 1,
\quad
z = \frac{1}{4} \sum_{i=1}^{4} \tilde{r}_i.
\]

This reward reflects the overall summary quality and guides the tree search toward higher-quality summary combinations.

\begin{figure}[t]
\centering
\includegraphics[width=0.44\textwidth]{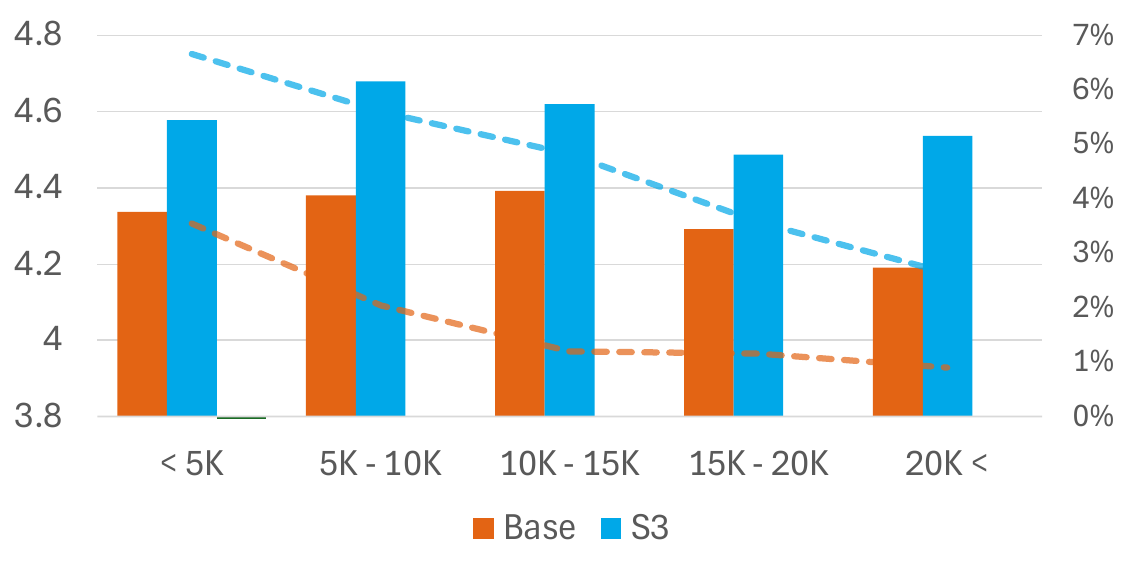}
\caption{Performance comparison across five length bins using Qwen2.5-7B. 
Bars represent \textit{relevance} scores, while dashed lines indicate summary length ratios (\%). The performance gap widens as input length increases, with S3 remaining robust on longer inputs.
}
\label{fig: length_bin}
\end{figure}

% S3 consistently achieves the best performance, maintaining strong results even in Bin5 with the longest inputs. Numeric labels denote summary length ratios (\%) relative to the source document.

\subsubsection{Backpropagation}
After each simulation, we update the statistics of all nodes along the selected search path. Specifically, for each visited state--action pair $(s_t, a_t)$, we increment the visit count and accumulate the reward $z$ obtained from the self-evaluation step:

\[
N(s_t, a_t) \leftarrow N(s_t, a_t) + 1,
\]
\[
W(s_t, a_t) \leftarrow W(s_t, a_t) + z.
\]
\[
Q(s_t, a_t) = \frac{W(s_t, a_t)}{N(s_t, a_t)}.
\]

During backpropagation, the value estimates of segment-level summary combinations along the search path are updated based on the obtained reward. Over multiple simulations, actions that consistently yield higher-quality summaries accumulate larger Q values, guiding the search toward actions with higher estimated value.

\subsection{Refinement}

After completing the simulations, we construct the draft summary by traversing the search tree from the root to a leaf, selecting the child with the highest Q value at each depth. The resulting segment-level summaries are concatenated to represent the entire meeting. We then apply a refinement stage to improve global coherence and reduce redundancy. Since LLMs often begin with generic high-level phrases (e.g., ``This document discusses ...'') regardless of the segment depth, such expressions are unnecessary for intermediate segments. As a result, directly concatenating segment-level summaries may introduce redundant discourse markers and weaken structural coherence. The refinement stage addresses these issues, producing a non-redundant and well-structured final summary.

% To mitigate this issue, we apply a refinement step after selecting and concatenating segment-level summaries. The refinement process removes redundant expressions across segments, eliminates unnecessary generic phrases, and improves overall coherence, producing a concise and well-structured final summary.

% tend to generate generic discourse-level phrases when generating summaries for individual document segments, LLMs often begin 

\section{Experimental Setup}

\subsection{Datasets}
% We conduct experiments on QMSum~\cite{zhong-etal-2021-qmsum}, a multi-domain long meeting summarization dataset. QMSum contains meeting transcripts from diverse sources, including ICSI~\cite{icsi} and AMI~\cite{ami} meeting corpora, as well as parliamentary proceedings such as the Welsh and Canadian Parliament. The dataset consists of long and complex meeting conversations, making it well-suited for evaluating long meeting document summarization systems.

We conduct experiments on QMSum~\cite{zhong-etal-2021-qmsum}, a multi-domain dataset for long meeting summarization. It includes transcripts from diverse sources such as the ICSI~\cite{icsi} and AMI~\cite{ami} meeting corpora, as well as parliamentary proceedings from the Welsh and Canadian Parliaments. Its length and complexity make it well-suited for evaluating long meeting summarization.

\subsection{Models}

We evaluate the effectiveness of S3 by comparing it against various LLM families and model sizes. In particular, we select models that are capable of processing long meeting documents from the QMSum dataset, including Qwen series (Qwen-2.5-7B-Instruct, Qwen-2.5-72B-Instruct)~\cite{qwen2.5}, and Gemma-3-12b-it~\cite{gemma_2025}. For each backbone model, all components of S3---including segment-level summarization, MCTS, and refinement---are performed using that same backbone. 
% This design ensures a consistent evaluation setting and isolates the effect of the S3 search framework from differences in model architectures or external evaluators.

We compare S3 with both summary-level and segment-level baselines. For summary-level baselines, we consider (1) a zero-shot model that generates the entire summary in a single pass (Base), and (2) a refinement-based variant that performs an additional refinement step on the generated summary (Base~+~Refine). For the segment-level baseline, (3) we generate a single summary segment for each document segment without sampling multiple candidates, followed by a refinement step; we denote this variant as S2.

% For the segment-level baseline, (3) we generate a single summary segment for each document segment—without sampling multiple candidates—and then apply a refinement step; we refer to this variant as S2.
% For comparision models, We also include a segment-level baseline (S2) that generates a single summary per segment followed by refinement, without performing tree search.

\subsection{Metrics}

We use G-Eval~\cite{liu-etal-2023-g} as our primary evaluation metric, as it correlates strongly with human judgment and does not rely on reference summaries. Specifically, G-Eval evaluates summaries along four dimensions—\textit{coherence}, \textit{consistency}, \textit{fluency}, and \textit{relevance}—capturing key error types such as missing information and hallucinations.
Reference-based metrics such as ROUGE~\cite{lin-2004-rouge} and BERTScore~\cite{zhang2020bertscore} are widely used in summarization, but they primarily measure lexical overlap and embedding similarity with the reference summary. 
However, reference summaries in existing meeting datasets are typically short and highly compressed, which may penalize informative outputs and underestimate content coverage. As a result, these metrics may not adequately reflect the overall quality and comprehensiveness of long, information-rich meeting summaries.
Moreover, model-based evaluation frameworks such as QuestEval~\cite{scialom-etal-2021-questeval} and UniEval~\cite{zhong-etal-2022-towards} evaluate summaries by jointly considering the input document and the generated summary; however, they are typically constrained by input length limitations, making them unsuitable for long meeting transcripts.
% We therefore focus our evaluation on G-Eval, which better reflects summary quality in our setting.

\subsection{Hyperparameters}

In our experiments, we partition each document using a sliding window with window size $w=2048$ tokens and stride $r=256$ tokens. We set the sampling size to $k=5$. The number of simulations is fixed at 30, and the exploration constant $c$ in MCTS is set to 1.0. For candidate generation, we consider two decoding strategies. In the nucleus sampling setting, we use $p=0.9$ with a temperature 1.1. In the DBS setting, we set the diversity penalty to 0.8. For S2, we use greedy decoding. A repetition penalty of 1.1 is applied in all generation settings.

\begin{table}[t]
\centering
\caption{Length statistics of summaries on QMSum (Test). Ratio denotes the average word count divided by the average document word count.}
\small
\scalebox{0.77}{
\begin{tabular}{l|ccccccc}
\toprule
\multirow{3}{*}{\textbf{Text}} 
& \multicolumn{3}{c}{\textbf{\# Sentences}} 
& \multicolumn{3}{c}{\textbf{\# Words}} 
& \textbf{Ratio} \\
\cmidrule(lr){2-4}\cmidrule(lr){5-7}\cmidrule(lr){8-8}
& Avg & Min & Max 
& Avg & Min & Max 
& (\%) \\
\midrule
Document & 734.3 & 265 & 1.6K & 10.4K & 2.3K & 23.9K & - \\
\midrule
Reference & 3.4 & 1 & 10  & 64.7  & 14 & 188 & 0.62 \\
Qwen2.5-7B & 10.2 & 1 & 37 & 162.9 & 24 & 448 & 1.56 \\
Qwen2.5-72B & 14.5 & 2 & 48 & 238.8 & 41 & 725 & 2.29 \\
S3-7B & 23.3 & 3 & 53 & 486.9 & 72 & 884 & 4.66 \\
\bottomrule
\end{tabular}
}
\label{tab: qmsum_length_statistics}
\end{table}

\section{Results}

\subsection{Main results}

% We compare summary-level and segment-level summarization strategies.
% For summary-level baselines, we consider (1) a zero-shot model that generates the entire summary in a single pass, and (2) a refinement-based variant that performs a refinement step on the generated summary. For the segment-level baseline, we consider (3) generating a single summary segment for each document segment, without generating multiple candidates, and then apply a refinement step (S2). 
As shown in Table~\ref{tab: main_result}, the experimental results show that S3 consistently outperforms all comparison models across LLM families. Notably, S3 with a 7B model even surpasses the 72B summary-level baseline, demonstrating the effectiveness of our search-based segment composition strategy. Furthermore, the segment-level S2 approach outperforms the summary-level refinement baseline. This suggests that although modern LLMs can process long inputs, generating summaries in smaller, localized units followed by refinement leads to better overall quality. These findings highlight the limitations of direct long-context summarization and underscore the importance of structured, segment-wise composition for long meeting summarization.

\subsection{Performance across input length bins}

To analyze model behavior under varying input lengths, we partition the test set into five bins based on source document length (in 5,000-token increments) and report the G-Eval \textit{relevance} scores (Figure~\ref{fig: length_bin}), which measure alignment with the key discussions of the meeting.
While the Base model degrades as input length increases, S3 remains stable, with the gap widening in the longest bin. For inputs over 10K tokens, the Base model generates summaries shorter than 1\% of the source document length, whereas S3 maintains substantially longer and more informative outputs.
These results suggest that although modern LLMs can handle long contexts, preserving discussion alignment becomes increasingly challenging as input length grows, particularly without structured long-context strategies.

\begin{table}[t]
\centering
\caption{Comparison of models by different decoding strategies.}
\scalebox{0.82}
{\begin{tabular}{l|ccccc}
\toprule
\textbf{Model} & \textbf{Coh.} & \textbf{Con.} & \textbf{Flu.} & \textbf{Rel.} & \textbf{Avg.} \\
% \midrule
% Llama-3.2-Instruct (3B) & 3.78 & 3.86 & 4.70 & 4.05 & 4.10 \\
% \quad w/ \textbf{S3 (DBS)} & 4.09 & \textbf{4.13} & \textbf{4.92} & 4.41 & 4.39  \\
% \quad w/ \textbf{S3 (Nucleus sampling)} & \textbf{4.10} & 4.12 & \textbf{4.92} & \textbf{4.47} & \textbf{4.40}  \\
\midrule
Qwen2.5-Instruct (7B) & 4.01 & 4.22 & 4.90 & 4.36 & 4.37  \\
\quad w/ \textbf{S3 (DBS)} & 4.15 & 4.36 & 4.93 & 4.60 & 4.51  \\
\quad w/ \textbf{S3 (Nucleus sampling)} & \textbf{4.22} & \textbf{4.43} & \textbf{4.94} & \textbf{4.65} & \textbf{4.56} \\
\bottomrule
\end{tabular}}
\label{tab: decoding_result}
\end{table}

\subsection{Summary length statistics}

% 버전1: 글쓰기 가이드라인을 참조한다.
% According to general writing practice, writing guidelines suggest that summaries often comprise approximately 5–10\% of the original text~\cite{}. As shown in Table~\ref{tab: qmsum_length_statistics}, the reference summaries in our dataset account for only 0.62\% of the source document length, indicating that they are extremely concise and likely omit substantial information. In contrast, S3 generates summaries whose average length corresponds to 4.66\% of the original document, approaching the commonly recommended range and providing substantially richer informational coverage. The 72B summary-level baseline generates longer summaries (2.29\%) compared to the 7B baseline (1.56\%), yet these remain considerably shorter than the suggested range and appear insufficient to capture the full scope of long meeting discussions.

% 버전2: 참조하지 않고 구체적인 5%-10% 숫자를 뺀다.
As shown in Table~\ref{tab: qmsum_length_statistics}, the reference summaries in QMSum account for only 0.62\% of the source document length, representing an extreme compression ratio for long-form meeting transcripts. Such a high compression rate likely limits content coverage and provides weak supervision for generating informative summaries. In contrast, S3 produces summaries whose average length corresponds to 4.66\% of the original document, substantially increasing informational coverage while remaining concise. Although the 72B summary-level baseline generates longer summaries (2.29\%) than the 7B baseline (1.56\%), both remain far shorter than S3 and appear insufficient to capture the full scope of long meeting discussions.

\subsection{Candidate generation strategies}

We measure the performance of S3 under different candidate generation strategies (Table~\ref{tab: decoding_result}). We observe that nucleus sampling consistently outperforms DBS in the segment-level tree search setting. We attribute this to the greater structural diversity induced by stochastic sampling, which allows the tree search to explore a broader combination space. In contrast, DBS tends to generate high-probability yet structurally similar candidates, which limits the effectiveness of subsequent search.

\section{Conclusion}

In this paper, we proposed S3, a training-free segment-level summarization framework based on Monte Carlo Tree Search for long meeting documents. By exploring combinations of segment-level summaries through structured search, S3 enables an effective global composition of dispersed information. Experiments show that S3 consistently outperforms strong summary-level and segment-level baselines across multiple LLM families, highlighting the importance of structured composition over model scale or context length alone.

\section{Acknowledgments}

This work was supported by the IITP(Institute of Information \& Coummunications Technology Planning \& Evaluation)-ITRC(Information Technology Research Center) grant funded by the Korea government(Ministry of Science and ICT)(No. IITP-2026-RS-2024-00437866) (45\%)
; by Culture, Sports and Tourism R\&D Program through the Korea Creative Content Agency grant funded by the Ministry of Culture, Sports and Tourism in 2025 (No. RS-2025-02413038, Development of an AI-Based Korean Diagnostic System for Efficient Korean Speaking Learning by Foreigners) (45\%)
; and by Institute of Information \& communications Technology Planning \& Evaluation (IITP) grant funded by the Korea government(MSIT) (No. RS-2019-II191906, Artificial Intelligence Graduate School Program (POSTECH)) (10\%).

\bibliographystyle{IEEEtran}
\bibliography{mybib}

\end{document}